\documentclass[a4paper]{ms}

\usepackage{graphicx}
\usepackage{amssymb,amsmath,array, amsfonts}
\usepackage[latin1]{inputenc}
\usepackage[natbib,backend=biber,sorting=none,maxbibnames=99, giveninits=true]{biblatex}
\bibliography{ms.bib}

\usepackage{tikz} %
\usepackage{tikz-3dplot}
\usepackage{pgfplots}
\usepackage{subcaption}

\pgfplotsset{compat=1.5}
\pgfplotsset{samples=200}
\usepackage{lipsum}
\usepackage{booktabs}
\usepackage{multirow}
\usepackage{float}
\usepackage{needspace}

\usepackage{color}
\usetikzlibrary{spy}
\usetikzlibrary{patterns}
\usepackage{siunitx,calc}

\DeclareCaptionFont{white}{\color{white}}
\newcommand{\mycite}[1]{\citeauthor{#1}~\cite{#1}}

\DeclareCaptionFont{white}{\color{white}}
\usepackage{placeins}
\usepackage{float}
\usepackage{eso-pic}

\AtBeginDocument{\AddToShipoutPictureFG*{\AtTextUpperLeft{\put(0,\LenToUnit{40pt}){\parbox{\textwidth}{\centering\bfseries
					27th European Symposium on Artificial Neural Networks, Computational Intelligence and Machine Learning (ESANN), Bruges, Belgium, 2019
}}}}}

\begin{document}
	\title{Learning Super-resolution 3D Segmentation of Plant Root MRI Images from Few Examples}
	
		\author{Ali Oguz Uzman, Jannis Horn and Sven Behnke
		\thanks{This research was supported by grant BE 2556/15 of German Research Foundation (DFG). We thank Andrea Schnepf (Bonn/FZJ) for providing MRI and annotations.}
		\vspace{.3cm}\\
		University of Bonn, Computer Science Institute VI, Autonomous Intelligent Systems \\
		Endenicher Allee 19a,
		53115 Bonn, Germany
	}
	
	\maketitle
	
	\begin{abstract}
		Analyzing  plant roots is crucial to understand  plant performance in different soil environments. While magnetic resonance imaging (MRI) can be used to  obtain 3D images of  plant roots, extracting the root structural model is challenging due to highly noisy soil environments and low-resolution of  MRI images. To improve both contrast and resolution, we adapt the state-of-the-art method RefineNet for 3D segmentation of the plant root MRI images in super-resolution. The networks are trained from few manual segmentations that are augmented by geometric transformations, realistic noise, and other variabilities.  The resulting segmentations contain most root structures, including branches not extracted by the human annotator.
	\end{abstract}
	
	\section{Introduction}
\begin{figure}[t]
		\centering
		\newcommand{\imgwidth}{.22\textwidth}
		\newcommand{\imgwidthh}{.12\textwidth}
		\begin{subfigure}{\imgwidth} \centering
			\includegraphics[width=\textwidth]{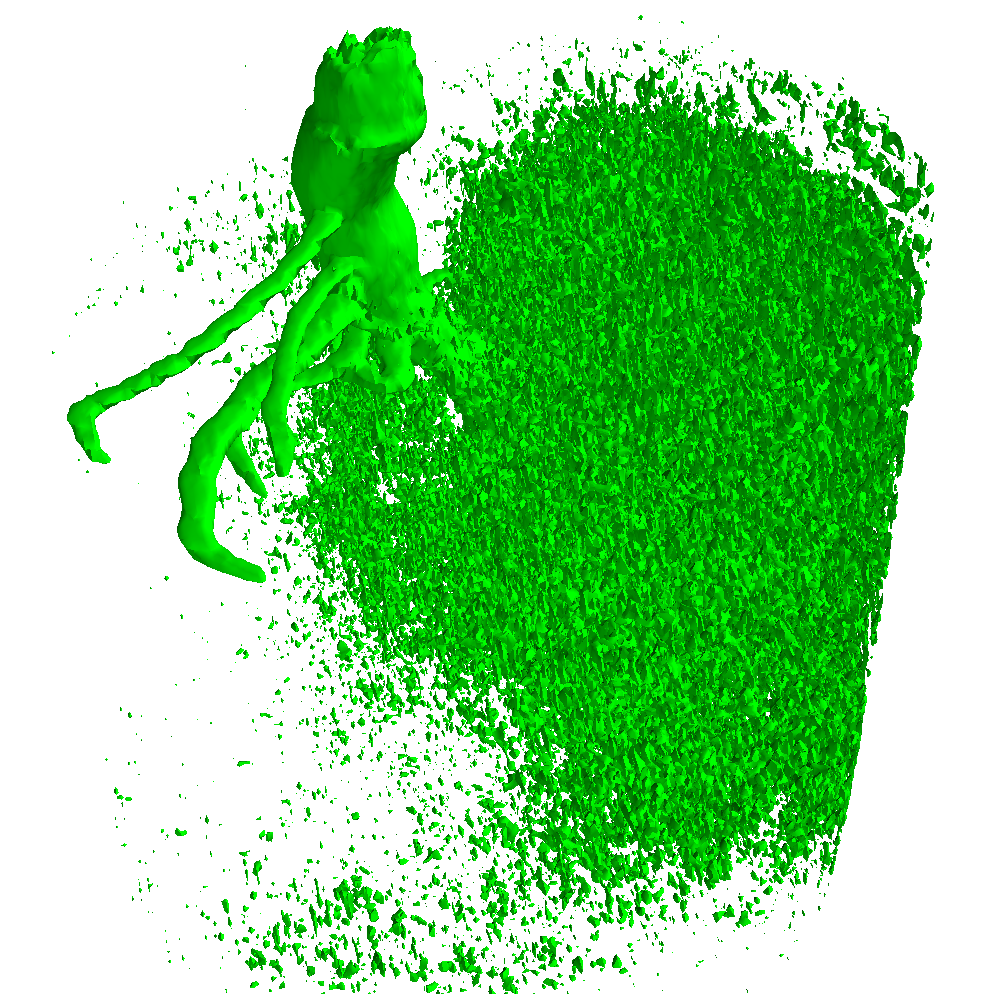}
		\end{subfigure}
		\begin{subfigure}{\imgwidth} \centering
			\includegraphics[width=\textwidth]{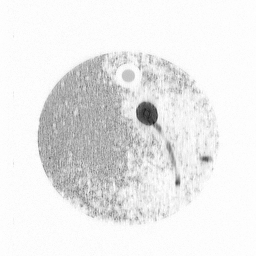}
		\end{subfigure}
		\begin{subfigure}{\imgwidth} \centering
			\includegraphics[width=\textwidth]{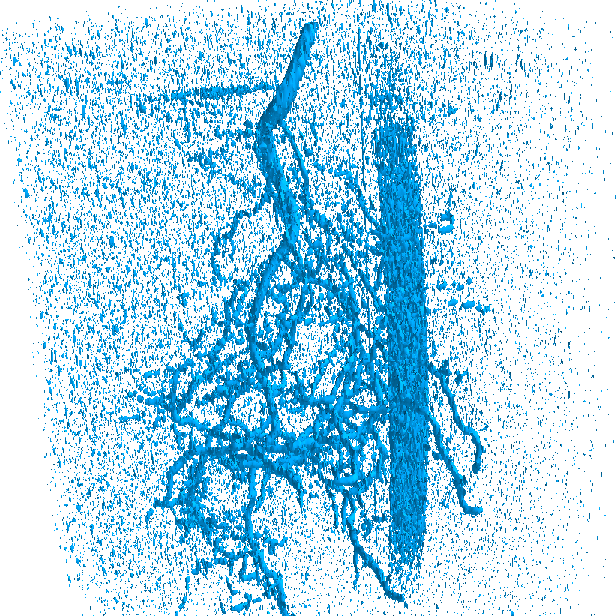}
		\end{subfigure}
		\begin{subfigure}{\imgwidth} \centering
			\begin{tikzpicture}[spy using outlines={circle,red,magnification=4,size=.5\textwidth, connect spies}]
			\node {{\pgfimage[interpolate=false,height=\textwidth]{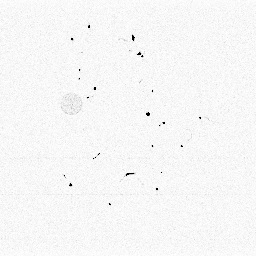}}};
			\spy[->] on (0.1,0.94) in node [left] at (1.4,-0.6);
			
			\end{tikzpicture}
		\end{subfigure}
		\caption{
			First two images depict the 3D visualization and a single slice from \textit{Root 1}. Third and fourth images depict the 3D visualization and a single slice (contrast enhanced \& magnified) from \textit{Root 2}.} \label{fig:noisyimages}
	\end{figure}

	Plant roots have been a long-standing research topic due to their crucial role for plants~\cite{nmrooting}. Their analysis requires extracting the root structural model. 
	To that end, 3D measurements of  plant roots in  opaque soil  are obtained with  MRI~(Figure~\ref{fig:noisyimages}) for use in root model extraction algorithms~\cite{nmrooting, reconroot, stingaciu2013situ}. 
	Often, the soil is noisy and the resolution of MRI  images is too low to capture thin roots with precision, resulting in low signal-to-noise ratio (SNR). With low SNR, automated extraction of root structural models is challenging~\cite{reconroot}, 
	requiring a preprocessing step to reduce the noise and increase the resolution artificially.
	In recent years, deep learning methods have shown great success with many computer vision tasks such as image classification~\cite{resnet},  action detection~\cite{actiondetection}, and semantic segmentation~\cite{refinenet}. 
	To improve both contrast and resolution of MRI, we adapt the state-of-the-art transfer learning method  RefineNet~\cite{refinenet} for 3D, \mbox{super-resolution} segmentation of plant root MRI images as root vs non-root. Since the original data we were provided with is insufficient for training, we generate our own synthetic training data from reconstructed root structures.
	
	\section{Related Work}
	Recent deep learning methods  define the state-of-the-art results for semantic segmentation~\cite{refinenet, deeplabv3}. Using a technique known as transfer learning, the features learned from large data sets  are leveraged to initialize training on a small, related data set, resulting in shorter training times and less need for training data.
	
	Despite extensive work on  semantic segmentation for 3D medical data, the use of 3D CNNs remains limited due to high memory and computational power requirements of training. 
	For example, a 3D CNN is proposed by~\mycite{kleesiek2016deep} for skull stripping from MRI images. However, the dataset is rather small and the resulting segmentations have lower resolution than the network input. 
	
	2D CNNs can also be utilized for segmentation of 3D data.
	A 2D CNN by \mycite{brainmri3dseg}  achieves comparable results to the state-of-the-art results at brain tumor segmentation from MRI images.
	
	In addition to semantic segmentation, deep learning methods also define  state-of-the-art results for super-resolution. Usually, the mapping from low-resolution to a higher resolution is done by incorporating upsampling~\cite{pham2017multi}, transposed convolutional layers~\cite{gansuper}, or serialization~\cite{behnke2001learning}.
	
	\mycite{nmrooting,reconroot} introduce algorithms for automated  structural model extraction from MRI root images, yet, these algorithms do not perform well on MRI images with low SNR. Using 3D CNNs, an attempt to increase the SNR by 3D segmentation of MRI images  has been made by \mycite{jannis_thesis}. While accurate results can be obtained,  higher memory consumption and longer training times prevent the construction of deep architectures. Moreover, training is difficult as the networks are highly volatile to parameterizations of learning rate, kernel size, number of channels, and number of layers. 
	\vspace{-0.05cm}
	\section{Segmentation Method}
	\vspace{-0.05cm}
	To increase the resolution, we target super-resolution factor $k=2$, i.e., mapping from an input image $I \in \mathbb{R}^{x\times y\times z}$ to a binary segmentation $S \in B^{2x \times 2y\times 2z}$. The non-root and root voxels are denoted with 0 and 1, respectively. We label the axis that follows the plant root from top to the bottom as $z$ while the other two axes are labeled with $x$ and $y$. A \textit{layer} is referred to a single slice of 1 voxel thickness on the $z$ axis.

	It has been reported that 2D CNNs can generate accurate segmentations of 3D data~\cite{brainmri3dseg}. Encouraged by the benefits of transfer learning~\cite{yosinski2014} and state-of-the-art results of the  RefineNet~\cite{refinenet}, we adapt 2D RefineNet for 3D segmentation of plant root MRI images in super-resolution. We apply  2D segmentation on each layer of the MRI image and obtain two consecutive segmented layers on the $z$ axis that have twice the resolution of the input in both $x$ and $y$ axes.
	
	\subsection{RefineNet}
\begin{figure}
		\centering
		\includegraphics[width=.9\textwidth]{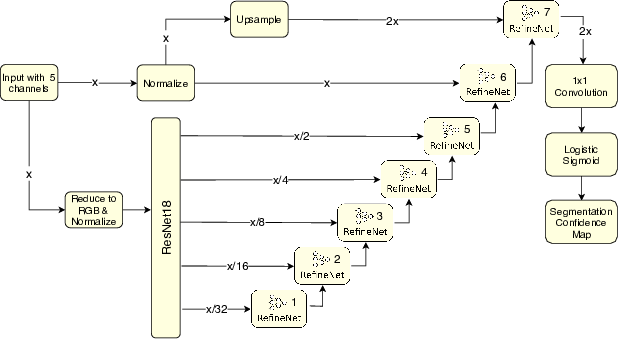}
		\caption{7-Cascade RefineNet. Includes figures from~\mycite{refinenet}.} \label{fig:7cascade}
		
	\end{figure}

	RefineNet~\cite{refinenet} exploits both high and low-level features of the pretrained networks by introducing RefineNet blocks. Through these blocks, feature maps with different resolutions are fused together to produce high-resolution outputs.
	\mycite{refinenet} cascade RefineNet blocks for best performance.
	We extend the original 4-Cascade RefineNet architecture~\cite{refinenet} to construct 7-Cascade RefineNet (Figure~\ref{fig:7cascade}) for image segmentation.

	We apply the segmentations layer-wise.  Since 3D information must also be exploited, for segmentation of each single layer, we use the layer itself and its four neighboring layers.
	Extracting features from ResNet requires 2D RGB images, thus, these five layers must be mapped to three channels. To compress the data into three channels while retaining maximum information, PCA is used. The first, second, and the third components of PCA are mapped to green, red, and blue respectively; according to the contribution of each channel into image luminance. 
	
	The cascaded RefineNet blocks are shown in Figure~\ref{fig:7cascade}. 
	As in the original 4-Cascade RefineNet~\cite{refinenet}, the resolution on both $x$ and $y$ axes are increased by a factor of two at each block. To achieve a super-resolution factor of two, we add three more blocks. Features that are extracted from five different stages of ResNet are fed to blocks  1--5.
	The remaining blocks 6\,\&\,7 operate on the original and super-resolution inputs. As they do not use the activations of ResNet, mapping to RGB is skipped; the original five layers of input are utilized directly by interpreting these layers as the channels of the network input.
	The 7th RefineNet block is followed by a $1\times1$ convolutional layer outputting two channels. Through the use of the logistic sigmoid function, the two channels are interpreted as two consecutive predicted layers where the voxel values denote the confidence of the network that the voxel belongs to a root.
	
	\subsection{Data Augmentation and Training}
	We were provided with four MRI root images and their manual root structure reconstruction. From the root structure, ground truth for the MRI images can be generated by voxelization. As the generated root voxels do not align well with the root structures in the MRI images, we cannot use real MRI images for training with the generated ground truth, though.
	We address this issue by generating artificial MRI images with perfectly aligned ground truth.
	Variety is introduced with transformations such as thickness adjustment, rotation, mirroring etc. 
	Modeled after the MRI images in Fig.~\ref{fig:noisyimages}, we generate soil noise in  different intensity scales using Perlin noise, uniformly and normally distributed noise for artificial MRI images. We use 384 artificial MRI--ground truth pairs for training and 384 for validation.
	The network is trained for 100 epochs with learning rate 6e-4 and gradient clipping 0.01 on artificial training pairs.
	\subsection{Evaluation}
	Only a small fraction of the number of voxels actually contain root, thus,  F1-Score is used as a metric for robustness against class imbalance.  On augmented data, we calculate the average F1-Score on the whole validation set. Moreover, the overall F1-Score for a given SNR range is also calculated. 
	
	The provided manual root annotations often contain misalignments with the real MRI images. We introduce a Distance Tolerant F1-Score for robustness against such small differences. To this end, we define distance tolerant precision $p'$ and recall $r'$  as follows:
	\begin{align*}
	p' = \frac{\sum_{i, j, k=0}^{\hat{x},\hat{y}, \hat{z}}dilate(G,d)^{i,j,k} \cdot S^{i,j,k}}{\sum_{i, j, k=0}^{\hat{x},\hat{y},\hat{z}}S^{i,j,k}}\text{,}&&r' =  \frac{\sum_{i, j, k=0}^{\hat{x},\hat{y}, \hat{z}}{G}^{i,j,k} \cdot dilate({S},d)^{i,j,k}}{\sum_{i, j, k=0}^{\hat{x},\hat{y}, \hat{z}}G^{i,j,k}}
	\end{align*}
	where $G, S\in \mathbb{B}^{\hat{x},\hat{y}, \hat{z}}$ are ground truth and predicted segmentation, respectively. Given a distance tolerance $d$, $dilate(I, d)$ refers to a morphological dilation of binary image $I$ by $d$ voxels. Using $p'$ and $r'$, F1-Score is calculated as usual.

	\vspace{0.5cm}
	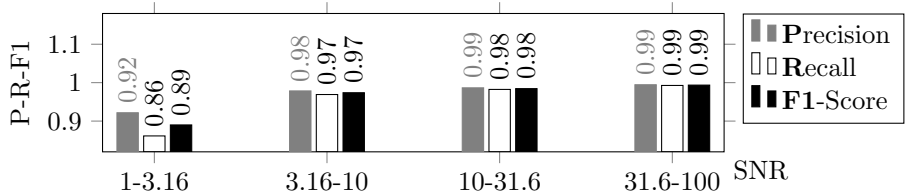
\begin{figure}[H]
		\centering
		\begin{tikzpicture}
		\pgfplotsset{width=.8\textwidth}
		\pgfplotsset{height=.28\textwidth}
		\begin{axis}[
		ybar,
		ymin = 0.82,
		enlargelimits=0.1,
		legend style={at={(0.5,-0.15)},
			anchor=north,legend columns=1},
		ylabel={P-R-F1},
		symbolic x coords={1-3.16,3.16-10,10-31.6,31.6-100, Average},
		xtick={1-3.16,3.16-10,10-31.6,31.6-100},
		nodes near coords,
		ytick align=outside,
		nodes near coords align={vertical},
		legend cell align={left},
		xlabel=SNR,
		x label style={at={(axis description cs:1,0)},anchor=north west},
		bar width=8pt,
		enlarge y limits={upper,value=1},
		legend pos =  outer north east,
		every node near coord/.append style={rotate=90, anchor=west},
		ymax=1.,
		]
		
		\addplot [pattern=crosshatch, color=gray]coordinates {(1-3.16,0.9216839093494303) (3.16-10,0.9785010264312451) (10-31.6,0.9865158001079964) (31.6-100,0.9945621404121744) (Average,)};
		\addplot [] coordinates {(1-3.16,0.8613961476585938) (3.16-10,0.9691310948606312) (10-31.6,0.9824113488778395) (31.6-100,0.9927393341012372) (Average,)};
		\addplot [pattern=crosshatch, color=black] coordinates {(1-3.16,.89) (3.16-10,0.9737935215799929) (10-31.6,0.9844592963966184) (31.6-100,0.9936499012932423) (Average,) };
		\legend{\textbf{P}recision, \textbf{R}ecall, \textbf{F1}-Score}
		\end{axis}
		\end{tikzpicture}
		\caption{Segmentation performance on the augmented data validation set.}
		\label{fig:aug_f1}
	\end{figure}
	\section{Results}
\begin{figure}[H]
		\pgfplotsset{width=.7\textwidth}
		\pgfplotsset{height=.33\textwidth}
		\centering
		\begin{tikzpicture}
		\begin{axis}[xlabel=Distance tolerance in voxels,ylabel=P-R-F1,legend pos = outer north east, legend cell align={left}, ymin= -0.1, ymax=1.1,legend style={font=\fontsize{8}{1}\selectfont}]
		,
		\addplot[color=red, mark=+, mark options={scale=1}] table [skip first n=1, y index=1] {real_small.vals};
		
		\addplot[color=red, mark=x, mark options={scale=1}] table [skip first n=1, y index=2] {real_small.vals};
		
		\addplot[color=red, mark=*, mark options={scale=1}] table [skip first n=1, y index=3] {real_small.vals};

		\addplot[color=blue, mark=+, mark options={scale=1}, dashed] table [skip first n=1, y index=1] {real_22.vals};
		
		\addplot[color=blue, mark=x, mark options={scale=1}, dashed] table [skip first n=1, y index=2] {real_22.vals};
		
		\addplot[color=blue, mark=*, mark options={scale=1}, dashed] table [skip first n=1, y index=3] {real_22.vals};

		\addlegendentry{Root 1 - \textbf{P}recision}
		\addlegendentry{Root 1 - \textbf{R}ecall}
		\addlegendentry{Root 1 - \textbf{F1}-Score}
		\addlegendentry{Root 2 - \textbf{P}recision}
		\addlegendentry{Root 2 - \textbf{R}ecall}
		\addlegendentry{Root 2 - \textbf{F1}-Score}

		\end{axis}
		\end{tikzpicture}

		{	\vspace{0.5cm}\begin{minipage}[c]{0.49\textwidth}
				\centering
				\sf Root~1\,\includegraphics[width=.64\textwidth]{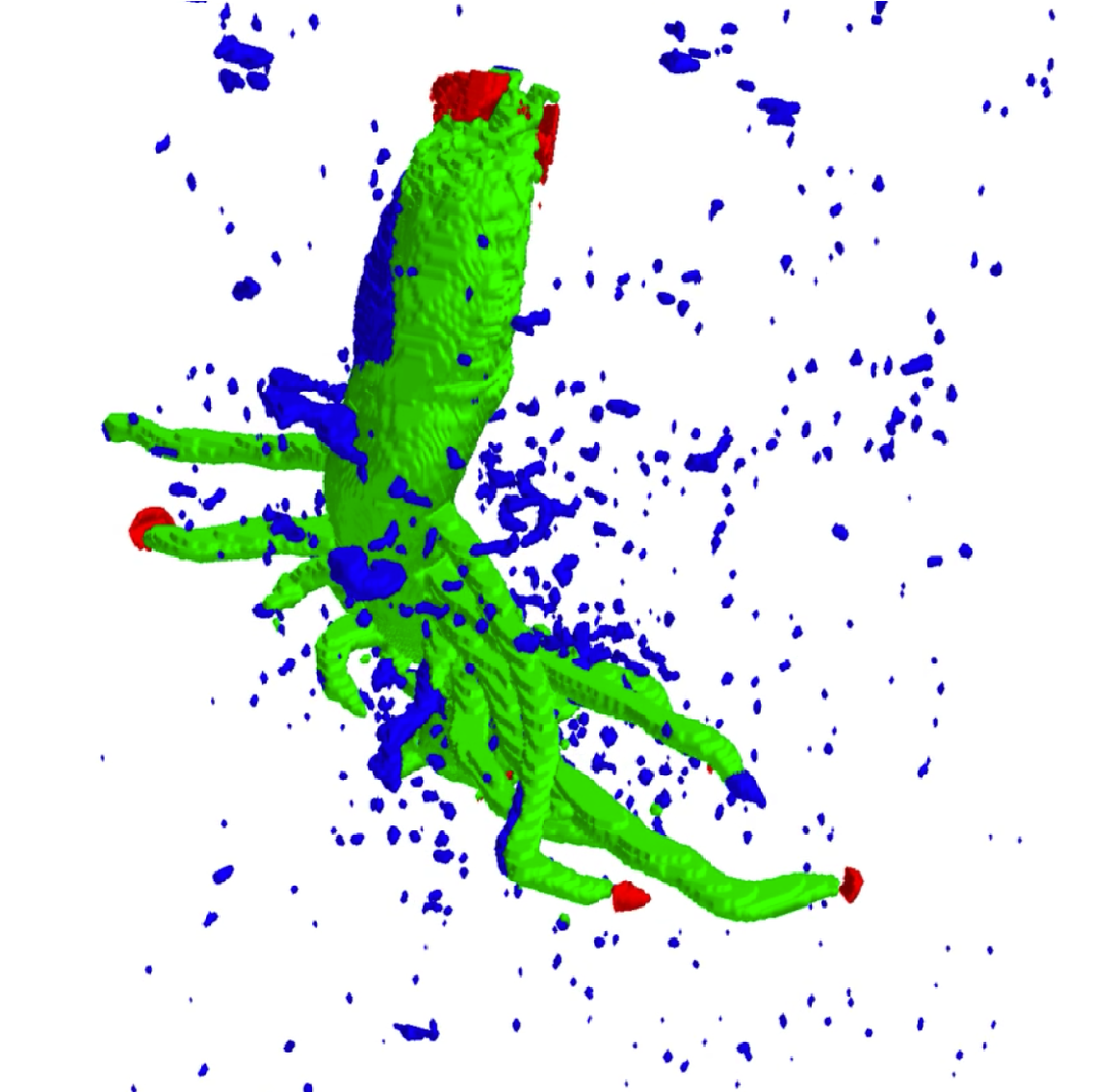}
		\end{minipage}}
		{
			\begin{minipage}[c]{0.49\textwidth}
				\centering
				\sf Root~2\,\includegraphics[width=.64\linewidth]{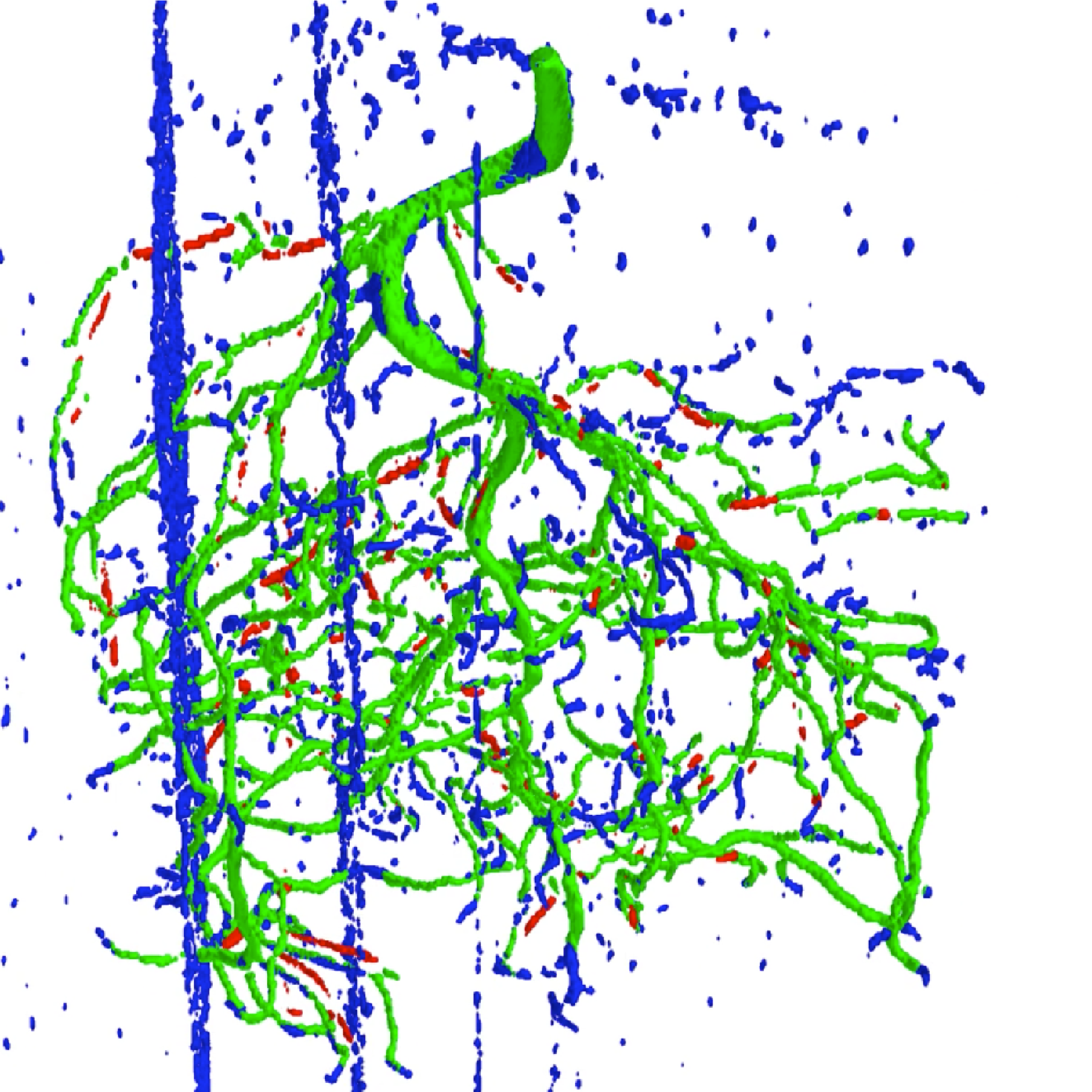}
		\end{minipage}}
		\caption{Performance on real MRI. The bottom row shows segmentations with distance tolerance of 5 voxels~(best viewed in color). Green, red and blue represent true positives, false negatives, and false positives, respectively.  \mbox{\textit{Root 1}} contains  minor false positives and almost no false negatives. \textit{Root 2}  contains minor disconnectivities and MRI artifacts as false positive.}
		\label{fig:seg_out}
	\end{figure}

	We evaluate our method on both synthetic (validation set) and real (test set) MRI data. The results on the validation set are presented in Fig.~\ref{fig:aug_f1}. Compared to the mean F1-Score 0.84 of \mycite{jannis_thesis}, ours is much better: 0.95. We further train a RefineNet architecture with the transfer learning layers removed; the average F1-Score drops from 0.95 to 0.91. For SNR lower than 3.16, the average F1-Score is 0.84 compared to 0.89 of the original; the remaining SNR ranges do not show much difference. 
	The network is further tested on two real
	root MRI images using distance tolerant F1-Score (Figure~\ref{fig:seg_out}). The results indicate that most roots are reconstructed correctly. The super-resolution contours of the segmented roots are detailed and smooth. Our method even detects additional roots not reconstructed by the human annotator, which are counted as false positives. For \textit{Root 2}, our method is sensitive to elongated MRI artifacts, which have not been present in the synthetic training data.

	\section{Conclusion}
	\vspace{-0.05cm}
	We developed a 2D CNN for 3D segmentation of plant root MRI images in super-resolution. As the data we were provided with was too small to train a neural network, we used a transfer learning architecture and generated synthetic training data.  With the exception of some extreme cases where the roots are extremely thin, our network performs well on the augmented data. On real data, we successfully detect most of the root structures; including some that are missing in the manual annotations. MRI artifacts introduce false positives in our predictions, which could be mitigated by including these in data augmentation. In future work, the utility of root-thickness dependent losses, data from different growth stages, and additional MRI channels might be investigated.
	
	\AtNextBibliography{\footnotesize}
	\printbibliography
\end{document}